\documentclass[journal]{IEEEtran}

\usepackage{url}

\usepackage{graphicx}
\usepackage{amsmath,amsfonts}
\usepackage{amssymb}
\usepackage{multirow}
\usepackage{makecell}
\usepackage{arydshln}
\usepackage{url}
\usepackage{amsfonts}
\usepackage{amssymb} 

\usepackage{cite}

\usepackage{xcolor}

\begin{document}

\title{Graph Convolutional Network for Multi-Target Multi-Camera Vehicle Tracking}

\author{Elena Luna, Juan C. SanMiguel, Jos\'{e} M. Mart\'{i}nez, and Marcos Escudero-Viñolo
\thanks{This paragraph of the first footnote will contain the date on which you submitted your paper for review. }
\thanks{Elena Luna, Juan C. SanMiguel, Jos\'{e} M. Mart\'{i}nez, and Marcos Escudero-Viñolo are with the Video Processing and Understanding Lab, Universidad Aut\'{o}noma de Madrid, Spain, e-mail: \{elena.luna;~juancarlos.sanmiguel;~josem.martinez;~marcos.escudero\}@uam.es.}
}
\markboth{}
{Shell \MakeLowercase{\textit{et al.}}: Bare Demo of IEEEtran.cls for IEEE Journals}
\maketitle

\begin{abstract}
This letter focuses on the task of Multi-Target Multi-Camera vehicle tracking. We propose to associate single-camera trajectories into multi-camera global trajectories by training a Graph Convolutional Network. Our approach simultaneously processes all cameras providing a global solution, and it is also robust to large cameras' unsynchronizations. Furthermore, we design a new loss function to deal with class imbalance. Our proposal outperforms the related work showing better generalization and without requiring ad-hoc manual annotations or thresholds,  unlike compared approaches. Our code is available at
\url{http://www-vpu.eps.uam.es/publications/gnn_mtmc}.
\end{abstract}


\IEEEpeerreviewmaketitle

\section{Introduction}

% Definition MTMC vehicle tracking
\IEEEPARstart{M}{ulti}-Target Multi-Camera (MTMC) vehicle tracking is
essential for applications of video-based Intelligent Transportation Systems. 
MTMC tracking aims to localize and identify multiple targets throughout videos recorded by several cameras. The lack of appropriate publicly available datasets hindered the research in the task, however, the interest in it has grown since the release of the MTMC CityFlow benchmark \cite{tang2019cityflow}.

% Those applications include, among others, vehicle detection and classification, plates' detection, crowd behavior analysis, anomaly detection, parking spot occupancy detection and visual traffic monitoring.
% The most important advantages of video-based
% systems are: relatively low installation cost, lack of need for complex construction work during installation, and ability to perform wide spectrum of tasks (e.g., detection, classification, tracking,
% traffic law violations monitoring) in various environments and weather conditions [4]

% \textcolor{red}{[Disadvantage: fitted to the scenario]} 

The lack of diverse benchmarks also entails that top approaches are strongly adapted to the scarce validation scenarios. The proposals in \cite{chen2019multi,hsu2019multi,ye2021robust,yang2021tracklet,ren2021multi,liu2021city} manually define both entry/exit areas and motion patterns of the vehicles. This set of 
a-priori assumptions may not be correct nor generalizable to unseen scenarios. Other proposals \cite{hsu2019multi,Hou2019,he2019multi,qian2020electricity,chang2019ai,ye2021robust,li2021multi,yang2021tracklet,ren2021multi,shim2021multi,chen2019multi,wu2019multiview,tan2019multi} compute ad-hoc distance thresholds (such as spatial, temporal and/or between appearance features), whose values are heuristically set using validation data. Albeit useful to filter undesired vehicles and false trajectories, these strategies exhibit major limitations for unseen data. Instead of handling all the cameras at once, the common tren in MTMC tracking \cite{ye2021robust,ren2021multi,tan2019multi} is to address the multi-camera problem by solving a sequence of de-coupled pairwise matching sub-tasks, i.e., analyzing pairs of cameras independently. This sequential processing may yield local optimum solutions rather than a global one \cite{phillips2019all}.

\begin{figure*}
\centerline{\includegraphics[width=0.9\textwidth]{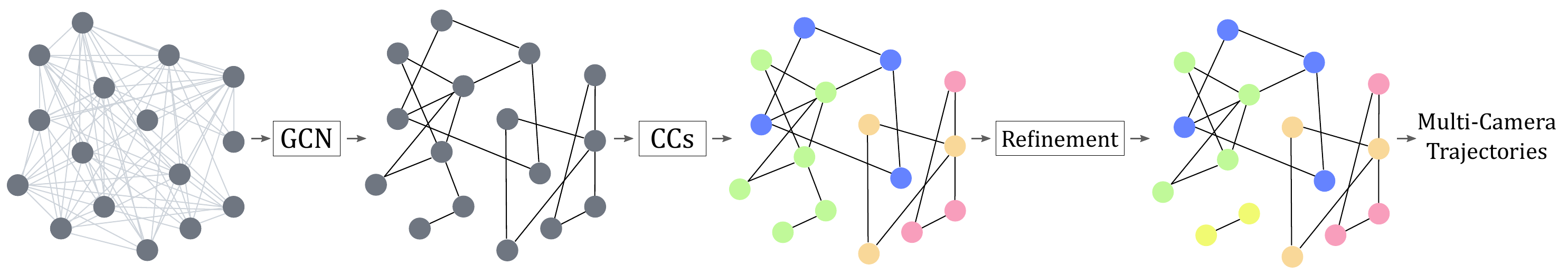}}
\caption{\textcolor{black}{The densely connected graph modeling all the possible associations is pruned by the GCN by edge classification. Connected Components (CCs) created on the resulting graphs are further refined to extract MC trajectories.  \vspace{-4 em} \label{fig:overview}}}
\end{figure*}

\begin{figure}
\centerline{\includegraphics[width=0.35\textwidth]{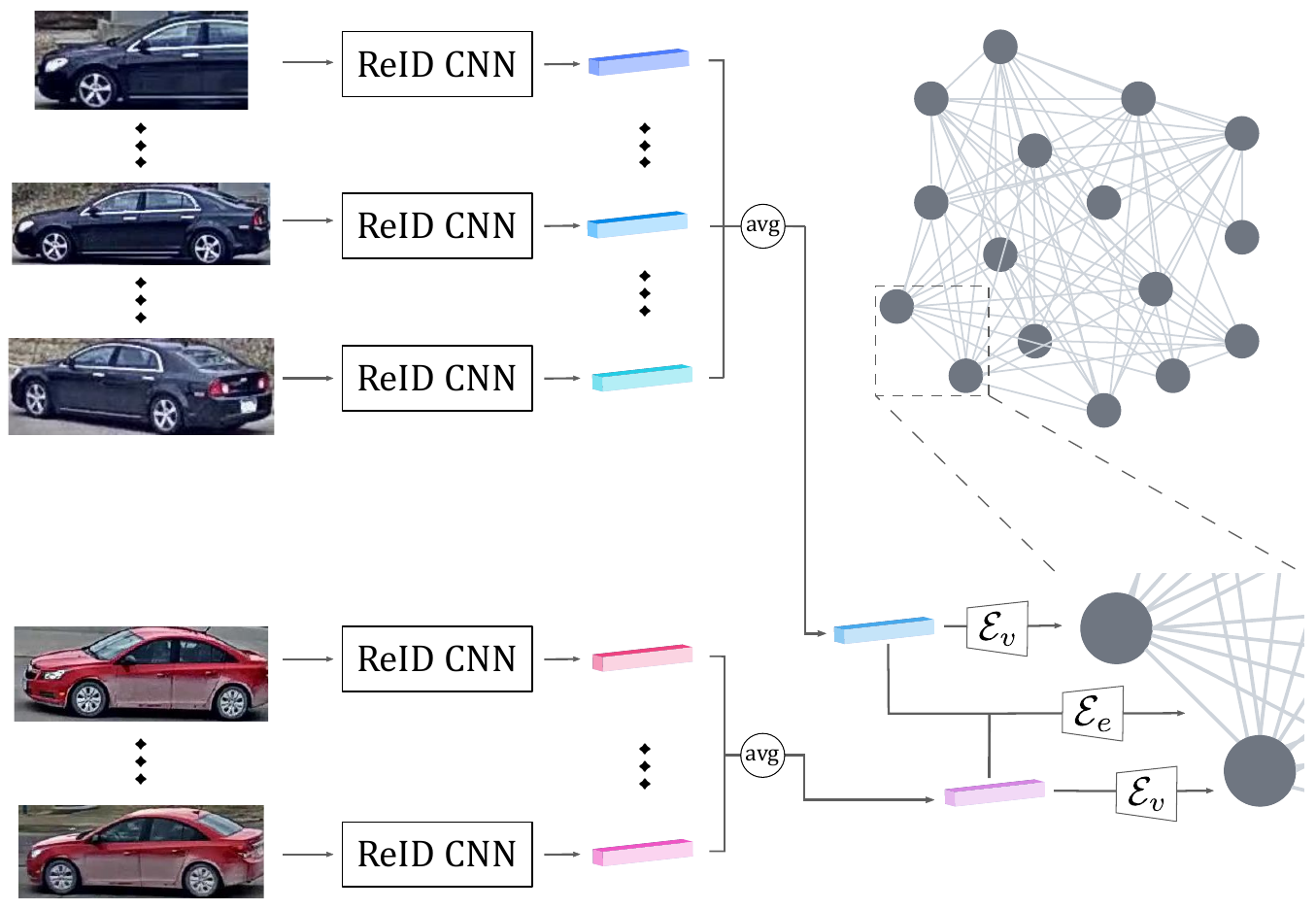}}
\caption{\textcolor{black}{Example of two nodes and edge initialization. \label{fig:graph_ini}}}
\end{figure}

% Cross-camera association challenge

The vast majority of the State of the Art (SOTA) proposals are offline  approaches \cite{ye2021robust,yang2021tracklet,ren2021multi,liu2021city,chen2019multi,li2021multi,shim2021multi,wu2019multiview,tan2019multi}, as global tracking optimization is prone to result in better performance than a frame-by-frame analysis. The offline proposals often follow a similar two-stage strategy: (1) computing Single Camera (SC) vehicle trajectories, and (2) performing cross-camera association to obtain the global Multi-Camera (MC) vehicle trajectories. Associating SC trajectories is a challenging task due to light changes, occlusions, and variations in vehicles' poses and appearances from different viewpoints.
For this reason, Re-Identification (ReID) features are widely used to get appearance embeddings  for estimating visual similarities. Some approaches \cite{li2021multi,ren2021multi,chen2019multi,hsu2019multi,Hou2019} associate trajectories based on thresholding such distances. Others find the optimal assignment by employing either the Hungarian algorithm \cite{shim2021multi,tan2019multi,ye2021robust}, or hierarchical clustering \cite{liu2021city,he2019multi}. Moreover, all these approaches apply ad-hoc constraints, such as spatial, temporal, direction, and velocity fixed thresholds to filter out trajectories associations. 

In another vein, some traditional graph-based approaches, where each SC trajectory is assigned to a node, have been proposed \cite{wu2019multiview,quach2021dyglip}.
% Associating SC trajectories is a challenging task due to light changes, occlusion patterns among the cameras, and, of course, due to the vehicles' pose and appearance variation in different viewpoints, i.e., distinct vehicles may appear quite similar from the same viewpoint, however the same vehicle from different viewpoints may be difficult to recognise.
An iterative Graph Matching Solver (GMS) was proposed at different levels  \cite{wu2019multiview}, where the graph is iteratively processed as a clique finding problem. Moreover, a dynamic graph can be created and expanded frame-by-frame \cite{quach2021dyglip}. To this aim, a multi-head Graph ATtention network (GAT), a type of Graph Neural Network (GNN), was trained to capture structural and temporal variations across time. 

% \textcolor{red}{[GNN related works]}
The main motivation behind GNNs is to extend Convolutional Neural Networks (CNNs), which have proven their effectiveness in regular Euclidean data, e.g., images, to a non-Euclidean domain, such as graphs. 
% In contrast to standard CNNs, the outputs of GNNs are invariant to the input order of the nodes \cite{zhou2020graph}. 
Furthermore, a core assumption of existing machine learning algorithms is that instances are independent of each other, which no longer holds for graph data \cite{wu2020comprehensive}. Thus, GNNs are able to model and exploit the relationships among the inputs, i.e., nodes.

% \textcolor{red}{[Imbalance ]}
% Imbalance data problems pose a difficulty for learning algorithms \cite{oksuz2020imbalance}. When they are not addressed, they have adverse effects on the final performance. Class imbalance occurs when the number of samples pertaining to different classes is significantly unequal. Existing algorithms for imbalanced learning can be divided into two groups, namely re-sampling and re-weighting approaches. Re-weighting approaches reweights the samples during training to reinforce the learning of minority classes. The weights are commonly defined by the inverse classes' frequency \cite{huang2016learning, wang2017learning} within the complete dataset, and then, they are applied to the computation of each mini-batch loss. As for classical deep learning networks, imbalance affects GNNs, although, for these networks, it has been substantially less explored than for their CNNs counterparts. \cite{liu2021pick} proposed PC-GNN to overcome the class imbalance for the node classification task by proposing a new sampling strategy. \cite{wang2021tackling} also tackles class imbalance by re-weighting the loss and proposing a new node-level index for the task of multi-class node classification. \cite{shi2020multi} proposed a conditional adversarial training to deal with imbalance for node classification and graph clustering tasks. 

Imbalance Data Problems (IDPs) are defined by long-tailed data distributions that may decrease the learning performance of low-populated classes \cite{oksuz2020imbalance,escudero2022ccl}.
To tackle IDPs, some approaches weight the samples during training to reinforce the learning of minority classes. These weights used in the mini-batch loss, are commonly defined by the inverse classes' frequency \cite{huang2016learning, wang2017learning} within the complete training set.
Although being subjected to similar effects, IDPs for GNNs have been seldom explored. The few strategies include alternative sampling for node classification \cite{liu2021pick}, loss re-weighting \cite{wang2021tackling}, and conditional adversarial training for graph clustering \cite{shi2020multi}.

% THE PROPOSAL This letter proposes an offline two-stage MTMC vehicle tracking approach based on GNNs that addresses the disadvantages described above. First, the single-camera vehicle trajectories are computed, and, second, the cross-camera association is performed. We train a spatial Graph Convolutional Network (GCN) to perform edge classification (connected/ not connected) over a graph where each node represents a SC trajectory, and edges stand for the relations between them. To handle class imbalance distributions, we propose a dynamic calculation of the weights at the mini-batch level\textemdash rather than at dataset level, as it is usually performed. Unlike many related works, our approach does not require any threshold search, and we do not manually pre-define any area in the camera views to perform motion pattern filtering. Thus, we propose a totally generalizable approach, that is not customized to the target scenario, and can be easily moved to different one.
% Furthermore, we process all the cameras at the same time, providing a global solution for the entire sequence at once.

To address all these challenges, this letter proposes an offline two-stages MTMC vehicle tracking approach based on GNNs. Firstly, the SC vehicle trajectories are computed, and, secondly, the cross-camera association is performed. We train a spatial Graph Convolutional Network (GCN), a type of GNN, to perform edge classification over a graph where each node represents a SC trajectory.
% Unlike many related works, we process all the cameras simultaneously, providing a global solution for the entire sequence. We do not pre-define any threshold, nor any area in the camera views to filter undesired trajectories. Thus, we propose a generalizable approach, not customized to the target scenario. To handle IDPs, we propose a new dynamically batch-wise weighted loss function.

The proposed approach is more versatile\textemdash not customized to the target scenario, than the SOTA as we: (1) provide a global solution for the entire sequence without imposing camera synchronization constraints, (2) do not set any threshold nor analysis area, and (3) handle IDPs by a novel dynamical batch-wise weighted loss function.

\section{Proposed Approach \label{sec:method}}

Given a scenario simultaneously recorded by $M$ cameras, let $\mathcal{O}=\left\{ o_i, \; i\in [1,O]\right\}$ be the set of vehicles in the scenario, where $o_i$ may be recorded by a single or multiple cameras, and $O$ the total number of distinct identities. Our proposal is intended for real scenarios, therefore, the synchronization of the cameras is not a requirement. \textcolor{black}{Overall, it aims to get a global trajectory for each existing vehicle by pruning a graph via edge classification with a GCN (Figure \ref{fig:overview})}.
% We define a two-stage framework: 1) Single-Camera (SC) trajectories generation and 2) global trajectories estimation.

\paragraph*{SC trajectories} SC trackers output sets of time-ordered bounding box images (BBs). For each SC trajectory, we feed the set of BBs to a ReID CNN, and then, average all the obtained feature embeddings to get a feature descriptor of the SC trajectory \textcolor{black}{(see Figure \ref{fig:graph_ini})}. 
Each SC trajectory is defined as $\mathcal{T}_{n,m}= \left\{\textbf{f}_n, c_{m}\right\}, n \in [1,N], m \in [1, M]$, being $c_m$ the camera, $\textbf{f}_n$ the feature descriptor, and $N$ the total number of SC trajectories.

\paragraph*{Cross-camera  association}
We train a GCN \cite{wu2020comprehensive} to determine the nodes' representations/states. For each node, a  graph convolutional layer encapsulates its hidden state by aggregating its own and neighbors' representations. If several layers are stacked, i.e., the aggregation is sequentially performed, each node gathers information from farther nodes. The final node state is used for classification. Since we perform edge classification, we extend this approach also to edge-level, so the final state of each edge is also computed by aggregating neighboring information.

\paragraph*{Graph definition and initialization}
Let us define a graph $G=(V,E)$, where each node $v_i \in V$ denotes each SC trajectory $\mathcal{T}_{n,m}$, and let $E$ be the set of edges connecting pairs of nodes as follows:
 \begin{equation} \label{Eq-1}
 E=\left\{\left(v_{i}, v_{j}\right), c_i \neq c_j, \; i,j \in [1,M] \right\}.
 \end{equation}
Note that nodes under the same camera are not connected among them.

All nodes and edges are initialized with their initial state embeddings. The initial node embedding is defined as $h_{v_{i}}^{0} = \mathcal{E}_{v}( \textbf{f}_i)$, where $\textbf{f}_i$ is the feature descriptor of the SC trajectory associated to node $v_i$ and $\mathcal{E}_{v}$ is a Multi Layer Perceptron (MLP) learnable encoder.

% \begin{equation}
%     h_{v_{i}}^{0} = \mathcal{E}_{v}( \textbf{f}_i), 
% \end{equation}
% where $\textbf{f}_i$ is the feature descriptor of the trajectory associated to node $v_i$ and $\mathcal{E}_{v}$ is a Multi Layer Perceptron (MLP) learnable encoder (FCs + ReLU). 

The initial edges embeddings encode the distance between the appearance descriptors of the nodes the edge is connecting. It is defined as:
\begin{equation} 
h_{(v_{i},v_{j})}^{0} = \mathcal{E}_{e}([\; ||\; \textbf{f}_i ,  \textbf{f}_j\;||_2,\;  cos(\textbf{f}_i , \textbf{f}_j )]),
\end{equation}
where $[\cdot,\cdot]$  means concatenation, $ cos $ stands for cosine distance and $\mathcal{E}_{e}$ is a MLP learnable encoder. Both encoders, $\mathcal{E}_v$ and $\mathcal{E}_e$, aim to learn optimal embeddings for the target association task. \textcolor{black}{See Figure \ref{fig:graph_ini}}.

\paragraph*{Graph convolutions} Once the initial embeddings are extracted, the information has to be propagated. Similarly to \cite{braso2020learning,luna2022graph}, each hidden edge state is computed as a combination of the initial states of the edge, and of the nodes that the edge is connecting, $h_{(v_{i},v_{j})} = \mathcal{U}_e ([ h_{v_i}^{0}, h_{v_j}^{0}, h_{(v_i,v_j)}^{0}])$, where $\mathcal{U}_e$ is a MLP learnable encoder. 

% \begin{equation}
%  h_{(v_{i},v_{j})} = \mathcal{U}_e ([ h_{v_i}^{0}, h_{v_j}^{0}, h_{(v_i,v_j)}^{0}]),
% \end{equation}
% where $\mathcal{U}_e$ is a MLP learnable encoder (FCs + ReLU). 

The choice of the node aggregation function may vary depending on the proposal \cite{zhou2020graph}. As defined in \cite{kipf2017}, we consider the summation as follows:
\begin{equation}
    h_{v_i} = \sum_{j\in \mathcal{N}(v_i)} m_{(v_i,v_j)},
\end{equation}
where $\mathcal{N}(\cdot)$ stands for the one-hop neighbourhood and $m_{(v_i,v_j)} = \mathcal{U}_v([ h_{v_i}^{0}, h_{(v_i,v_j)}])$, being $\mathcal{U}_v$ another MLP learnable encoder.

% where $\mathcal{N}(\cdot)$ stands for the neighbourhood and $m_{(v_i,v_j)}$ is defined as: 
% \begin{equation}
%     m_{(v_i,v_j)} = \mathcal{U}_v([ h_{v_i}^{0}, h_{(v_i,v_j)}]),
% \end{equation}
% being  $\mathcal{U}_v$ another MLP learnable encoder (FCs + ReLU).

Since $G$ is densely connected (see Equation 1), we stack a single convolutional layer, i.e., we perform the aggregation once. This helps to avoid overfitting, since with just one layer, all nodes share information among them.

\paragraph*{Classifier} 
Our final aim is to predict which edges in the graph are connecting nodes of the same vehicle identity (class 1), and which ones do not (class 0). For every edge $(v_i, v_j) \in E$, we compute the edge prediction $\hat{y}_{(v_i,v_j)}$ that provides the likelihood of the edge being connected, as $\hat{y}_{(v_i,v_j)} = \mathcal{C}(h_{(v_i,v_j)})$, where $\mathcal{C}$ is a MLP learnable classifier.

%  \begin{equation}
%       \hat{y}_{(v_i,v_j)} = \mathcal{C}(h_{(v_i,v_j)}),
%  \end{equation}
%  where $\mathcal{C}$ is a MLP learnable classifier.

\paragraph*{Loss function} 
We train our model to predict the following classes for each edge:
\begin{equation}
y_{(v_i, v_j)}=\left\{\begin{array}{ll}
1 & \exists \; o_{i} \in \mathcal{O} \; \textrm{ s.t. } \; v_i \in o_{i},  v_j \in o_{i} \\
0 & \textrm{ otherwise,}
\end{array}\right.
\end{equation}

Due to the intrinsic nature of $G$, and its high connectivity, this is a severe unbalanced problem, where $y_{(v_i, v_j)} = 1$ is a very minority class. Furthermore, due to the way we sample the data (see Section \ref{sec:abl-studies}), the inter-batch class imbalance distribution is quite variable. In order to strengthen the model to learn the minority class we perform a dynamic batch-wise sample weighting in the batch loss computation:
 
 \begin{equation}
 \mathcal{L}_{B}=\sum_{\left(v_{i}, v_{j}\right) \in E'}\frac{w_{y_{(v_i,v_j)}} \cdot\mathcal{L}_{CE}(\hat{y}_{(v_i,v_j)}, y_{(v_i,v_j)})}{\sum_{\left(v_{i}, v_{j}\right) \in E'}\; w_{y_{(v_i,v_j)}}},  
 \end{equation}
 where $E' \in E$ is the set of edges in a training mini-batch and $w_{y_{(v_i,v_j)}}$ is the class weight,  computed online at batch-level, according to the class frequency, as follows:

\begin{equation}
w_{y_{(v_i, v_j)}}=
\begin{cases}
\frac{1}{freq(0)}=\frac{n_0+n_1}{n_0}, & y_{(v_i, v_j)} = 0\\
\frac{1}{freq(1)}=\frac{n_0+n_1}{n_1}, & y_{(v_i, v_j)} = 1, 
\end{cases}
\end{equation}
 being $n_0$ and $n_1$ the number of samples, i.e., edges, from each class 0 and 1 in the batch. A dynamic batch-wise weight, although calculated differently, was also successfully applied in \cite{changbatch} to tackle the foreground-to-background space imbalance problem \cite{oksuz2020imbalance}.
 
 A False Positive (FP) predicted edge, may cause the merging of vehicles with different IDs. In order to explicitly penalize these mispredictions, we include a new term in the batch loss function:
  \begin{equation}
 \mathcal{L}_T= \mathcal{L}_B + FPR,\; \mathrm{being  }\; FPR=\frac{\mathrm{FP}}{\mathrm{FP}+\mathrm{TN}}. \label{Eq:loss}
 \end{equation}
TN stands for True Negative, i.e., edges correctly predicted as not connected. The FPR scores in [0,1]: the lower the less mispredictions.
% It is important to point out that, although we learn to classify two classes, we train our model as a multi-class problem with only two classes
 
\paragraph*{Inference and global trajectories generation} 
To obtain the global MC vehicle trajectories in the target scenario, we firstly extract the SC trajectories to create $G$ and, secondly, compute $\hat{y}_{(v_i,v_j)}$, as described previously, inferencing the previously trained GCN model. As we train a multi-class classification task, we avoid thresholding $\hat{y}_{(v_i,v_j)}$ and only conserve the edges whose probability for class 1 is higher, suppressing all the rest. Finally, by computing the Connected Components (CCs) over $G$, we group the nodes into clusters of SC trajectories. All the SC trajectories in the same component/cluster, compose a global MC vehicle trajectory. Since the number of cameras $M$ is known, it is reasonable to assume some constraints in the clusters computation, such as the maximum size of the cluster, which should be, at most, $M$ SC trajectories. To this aim, we follow the clusters refinement strategies defined in \cite{luna2022graph}.

 \section{Experiments}
\subsection{Evaluation Framework: Datasets and Metrics}
% \paragraph*{Dataset}
We use the CityFlow benchmark \cite{tang2019cityflow}, considering the training (S01, S03 and S04) and validation (S02) scenarios. S01, S03 and S04 comprise 184 IDs, 36 cameras and 32000 frames in total. S02 is composed of 4 cameras with partially overlapped fields-of-view (FoVs), consisting of 145 IDs along 8440 frames (2110 per camera). Note there is no data overlapping between the training and validation sets.

% \paragraph*{Metrics}
The CityFlow Ground-Truth (GT) contains the bounding boxes of all the vehicles seen at least by two cameras, labeled with global IDs. The CityFlow evaluation benchmark adopts the metrics: Identification Precision ($IDP$), Identification Recall ($IDR$), and F1 Score ($IDF_1$) \cite{ristani2016performance}. Since SC trajectories are not considered, the metrics are computed only over the MC trajectories. 

% \begin{equation}
% IDP=\frac{IDTP}{IDTP+IDFP},
% \end{equation}
% \begin{equation}
% IDR=\frac{IDTP}{IDTP+IDFN},
% \end{equation}
% \begin{equation}
% IDF_{1}=\frac{2 \cdot IDTP}{2\cdot IDTP+IDFP+IDFN},
% \end{equation}
% where $IDTP$, $IDFP$ and $IDFN$ stand for True Positive ID, False Positive ID and False Negative ID. These metrics score from 0 to 100, the higher the better. 

\begin{table}
\small
\begin{center}

\renewcommand{\arraystretch}{1.1}
\caption{\textcolor{black}{Specification of each learnable encoder, graph convolutional layer, and the classifier of the network.\vspace{-1 em} \label{tab:sizes-fc}}}
% \resizebox{0.4\textwidth}{!}{

\begin{tabular}{ccccc}
\hline 
 & Layer & Type & Input & Output \tabularnewline \hline
 \hline 
 % \multicolumn{5}{c}{ \textbf{Feature Encoders}} \tabularnewline
 % \hline 
\multirow{4}{*}{\rotatebox{90}{$\mathcal{E}_v$}} & 0  & FC + ReLU &  2048 & 1024\tabularnewline
& 1 & FC + ReLU &  1024 & 512\tabularnewline
& 2 & FC + ReLU &  512 & 128\tabularnewline
& 3
& FC + ReLU &  128 & 32\tabularnewline
\hdashline\multirow{2}{*}{\rotatebox{90}{$\mathcal{E}_e$}} & 0  & FC + ReLU &  2 & 4\tabularnewline
& 1 & FC + ReLU &  4 & 4\tabularnewline
\hdashline
% \multicolumn{5}{c}{\textbf{Graph Convolutional Layers}} \tabularnewline
 
\rotatebox{90}{$\mathcal{U}_v$} & 0  & FC + ReLU &  36 & 32 \tabularnewline
\hdashline\rotatebox{90}{$\mathcal{U}_e$} & 0  & FC + ReLU &  68 & 4 \tabularnewline
% \multicolumn{5}{c}{\textbf{Classifier}} \tabularnewline
\hdashline\rotatebox{90}{$\mathcal{C}$} & 0  & FC + Softmax &  4 & 2\tabularnewline

\hline

\end{tabular}
\end{center}
\end{table}

\subsection{Implementation Details}
Since we tackle MC tracking by associating SC trajectories, we rely on SOTA approaches for object detection, SC tracking and ReID feature extraction. Regarding object detection and SC tracking, to perform a fair comparison, we use the same combination as the top approach in CityFlow (SSD object detector \cite{liu2016ssd} + TNT tracker \cite{wang2019exploit}). We do not have any ReID requirement so we infer ReID 2048-d appearance vectors $\textbf{f}_n$ with the model provided by \cite{liu2021city} using ResNet-101 as backbone and trained using the training set of the CityFlow benchmark. \textcolor{black}{The GCN architecture (specified in Table \ref{tab:sizes-fc}) has 2.7M parameters being signficantly lower than that of the ReID model (44.76M) }. To reduce overfitting during training, we adopt color jitter, random erasing and, at SC trajectory level, random horizontal flip data augmentation. The initial learning rate is set to 0.01 and the Batch Size (BS) is set to 100. To minimize Equation \ref{Eq:loss} and optimize the network parameters, we adopt the SGD solver. To avoid training-from-scratch early instabilities, a linear warmup strategy \cite{goyal2017accurate} (5 epochs) is used. We train for 100 epochs decaying the loss at epoch 50. It has been implemented using Pytorch Geometric framework, running on a TITAN RTX 24GB GPU. \textcolor{black}{The inference of S02 takes $201.67\pm0.46$ secs. (59.89\% ReID CNN + 0.39\% GCN + 39.72\% refinement)}.

\subsection{Ablation studies \label{sec:abl-studies}}

\begin{table}

\begin{center}
\caption{\label{tab:BS}Ablation study on the Batch Size (BS), using GT SC trajectories as input.\vspace{-1 em} }
\tiny
\resizebox{0.35\textwidth}{!}{
\begin{tabular}{cccc}
\hline
BS & $IDP\uparrow$  & $IDR\uparrow$ & $IDF_1\uparrow$  \\
 \hline
\hline

16 & 79.51 & 11.37 & 19.90 \\
32 &  97.73 & 84.75 & 90.78 \\
64 &  99.07 & 98.67 & 98.87 \\
100 & 99.07 & 98.84 & 98.95 \\

\hline
\end{tabular}}
\end{center}
\end{table}

\textbf{Batch size.} In order to be independent on the selected detector-tracker pair, we employ GT SC trajectories as input. As the training data is sampled at ID level, BS=16 means that 16 vehicle IDs are considered to build $G$. For every ID, a node is created for each camera view, thus, although the BS is fixed, the graph contains between BS and $M \times $ BS nodes.  
% From Table \ref{tab:BS}, 16 seems to be too small for training our proposal. The poor performance is likely because such a small BS is causing the model to converge to a non-optimal minimum. 64 and 100 seem to be better values for our BS training. Larger BS are not considered, due to hardware constraints. However, the small increase between 64 and 100, suggests that only small gains in performance can be obtained by increasing BS. We choose BS=100 for further experiments.
In the light of Table \ref{tab:BS}, we discard BS=16 as it ends up in a local far-from-optimum minimum. BS of 64 and 100 yield both accurate and similar performances, discouraging the use of larger BS, that are, in any case, unattainable by our hardware setup.

\begin{table}

\begin{center}
\caption{\label{tab:Strategies-temp} Ablation study on the temporal threshold, in frames. Using as input SSD \cite{liu2016ssd} detector + TNT \cite{wang2019exploit} SC tracker. $|\cdot|$ stands for the cardinality of a set.\vspace{-1 em}}
\tiny
\resizebox{0.45\textwidth}{!}{
\begin{tabular}{ccccc}
\hline
Temporal th. & $|E|$ & $IDP\uparrow$  & $IDR\uparrow$ & $IDF_1\uparrow$  \\
 \hline\hline
 300 & 90,576  &  65.43 & 70.98  &  68.09  \\ 
 500 & 123,474  & 64.21 & 72.38  &  68.05  \\ 
 700 & 155,400 & 67.35 & 81.29  &  73.66  \\ 
 900 & 184,292   & 69.69 & 92.26 &  79.40 \\ 
 1100 & 236,068 & 70.92 & 92.02 & 80.10 \\
 \textbf{1300}  & \textbf{264,541}  & \textbf{73.04} & \textbf{92.05} & \textbf{81.45} \\
 1700 & 280,828 & 72.44 & 92.98 & 81.43  \\
2110 (full)   & 294,398  & 71.95 & 92.81 & 81.06 \\
\hline
\end{tabular}}

\end{center}
\end{table}

\textbf{Temporal thresholding.} As stated in Section \ref{sec:method}, for inference, $G$ considers all vehicles detected from all cameras along the whole sequence. It may seem reasonable that considering associations among vehicles' trajectories that are distant in time is unnecessary and computationally inefficient, and, thus, restricting the association temporarily would be appropriate. For this reason, we study the creation of a smaller temporally constrained graph that limits the possible associations among nodes. Table \ref{tab:Strategies-temp} shows the tracking performance with respect to different temporal thresholds. We can observe how the size of $G$ is reduced as the threshold is smaller, thus making its processing faster and lighter. However, the performance is drastically reduced when we temporally restrict by 300, 500, and even 700 frames (30, 50, and 70 secs.). The IDR reduction suggests that some associations are missed due to the temporal restriction. Since the cameras’ FoVs partially overlap, one can extrapolate that cameras in CityFlow dataset might not be temporally synchronized (which, in fact, is true). Due to this large unsynchronization, and to avoid ad-hoc parametrizations, we discard the temporal thresholding. However, it may be useful for reducing complexity in quasi-synchronized scenarios, and, also, for processing long sequences.

\textbf{Training loss.} Table \ref{tab:Strategies-loss} ablates the loss terms proposals. Due to the huge class imbalance, differently weighting the contribution to the loss of the samples of each class is an essential requirement. In absence of the weighting strategy, the trained model failed to learn the minority class, thus, an edge-less graph is created; since the Cityflow evaluation only takes into account MC trajectories, no results can be provided (see Table \ref{tab:Strategies-loss}). The FPR penalization improves the performance for all BS. We can observe that the enhancement is greater for smaller BS.

\begin{table}
\begin{center}
\caption{\label{tab:Strategies-loss} Ablation study on adding dynamic batch-wise weighting ($w$) and FPR to the final loss (\cite{liu2016ssd} detector + TNT \cite{wang2019exploit} SC tracker)}
\resizebox{0.4\textwidth}{!}{
\begin{tabular}{ccccccc}
\hline
BS   & FPR & $w$ & $IDP\uparrow$  & $IDR\uparrow$ & $IDF_1\uparrow$ & $\Delta\; IDF_1 $   \\
\hline\hline
16 &  & &-& - & - & -   \\ 
16 &  & \checkmark & 53.88 & 28.48 & 37.26 & -   \\ 
\textbf{16} & \checkmark & \checkmark& \textbf{54.76} & \textbf{34.06} &  \textbf{41.99} & + 12.69 \% \\ 
\hdashline
32 &  &  &-& - & - & -   \\ 
32 &  & \checkmark & 59.10 & 47.39 & 52.60 & -  \\ 
\textbf{32} & \checkmark & \checkmark & \textbf{67.42} & \textbf{59.80} & \textbf{63.38}  & + 20.49 \% \\ 
\hdashline
64 &  &  &-& - & - & -   \\ 
64 &  & \checkmark & 69.91 & 92.36 &  79.39 & -  \\ 
64 &  \checkmark & \checkmark &  \textbf{71.63} &  \textbf{91.10} &  \textbf{80.80} & + 1.05 \%   \\ 
\hdashline
100 &  & &-& - & - & -   \\ 
100 &  &  \checkmark & 71.35 & 92.08 & 80.40 & -  \\ 
\textbf{100} & \checkmark & \checkmark & \textbf{71.95} & \textbf{92.81}  & \textbf{81.06} &  + 0.82 \%\\ 

\hline
 \end{tabular}}

\end{center}
\end{table}

\subsection{Comparison with SOTA}
 Table \ref{tab:results-SoA} compares our proposal with related SOTA (ordered by $IDF_1$) in S02. Results show that our proposal outperforms the top SOTA approaches while not requiring any ad-hoc parametrization. This differentiates from SOTA approaches requiring specific thresholds for spatial distance, time, features and/or velocity \cite{hsu2019multi,Hou2019,he2019multi,qian2020electricity,chang2019ai,luna2022online}, manual definition of areas \cite{hsu2019multi} or assumptions about the vehicles' paths \cite{qian2020electricity}.

% Explicacion umbral IOU + Imagen ejemplo

\begin{table}
 \renewcommand{\arraystretch}{1.2}
\begin{center}
\caption{\label{tab:results-SoA}Comparison with the related SOTA. \vspace{-2 em} }
\resizebox{0.5\textwidth}{!}{
\begin{tabular}{cccccc}
\hline
& Object Detector & SCT & $IDP\uparrow$  & $IDR\uparrow$ & $IDF_1\uparrow$  \\
\hline\hline
Ours & SSD \cite{liu2016ssd} & TNT \cite{wang2019exploit} & 71.95 & \textbf{92.81} & \textbf{81.06} \\
UWIPL \cite{hsu2019multi}& SSD \cite{liu2016ssd} & TNT \cite{wang2019exploit} & 70.21 & 92.61 &79.87  \\
ANU  \cite{Hou2019}  & SSD \cite{liu2016ssd}  & custom \cite{Hou2019} & 67.53  & 81.99 & 74.06 \\
BUPT \cite{he2019multi} & FPN  \cite{lin2017feature} & custom \cite{he2019multi}  & \textbf{78.23} & 63.69 & 70.22\\
DyGLIP \cite{quach2021dyglip} & Mask-RCNN \cite{he2017mask} & DeepSORT \cite{wojke2017simple}  & -& - & 64.90 \\
Online-MTMC \cite{luna2022online} & EfficientDet \cite{tan2020efficientdet} & custom \cite{luna2022online}  & 55.15 & 76.98 & 64.26 \\
Online-MTMC \cite{luna2022online} & Mask-RCNN \cite{he2017mask} & custom \cite{luna2022online}  & 57.23 & 71.99 & 63.76 \\
ELECTRICITY \cite{qian2020electricity} & Mask-RCNN \cite{he2017mask} & DeepSORT \cite{wojke2017simple}  & - & - & 53.80 \\
NCCU \cite{chang2019ai}& FPN \cite{lin2017feature} & DaSiamRPN \cite{zhu2018distractor} & 48.91 & 43.35 & 45.97 \\
\hline
\end{tabular}}
\end{center}
\end{table}

\section{Conclusion}
The MTMC vehicle tracking SOTA is heavily guided by the CityFlow benchmark. Usually, the proposals in this scope are fully adapted, even tailored, to their scenarios. Due to all the types of ad-hoc parameterizations, these approaches cannot be extrapolated to other scenarios.  We firmly believe in making algorithms that are versatile for real setups, thus, we propose a not-parameterized approach, also robust to highly unsynchronized video streams. We process all the camera views at the same time, being robust to possible errors induced by pairwise processing, and we outperform the related SOTA.

% --------------------------------------------

% relying on the appearance descriptor 

% .............................

% s 

% \section*{References and Footnotes}

% \section*{References}
\bibliographystyle{IEEEtran}
\bibliography{main.bib}

\end{document}